\documentclass{article} 
\usepackage{iclr2019_conference,times}

\usepackage{hyperref}
\usepackage{url}

\usepackage[T1]{fontenc}
\usepackage[latin1]{inputenc}
\usepackage{subeqnarray}
\usepackage{amsfonts}
\usepackage{amsmath}
\usepackage{amssymb}
\usepackage{graphicx}
\usepackage{color}
\usepackage{stfloats}
\usepackage{balance}
\usepackage{epstopdf}
\usepackage{minibox}
\usepackage{xcolor,colortbl}
\usepackage{algpseudocode}
\usepackage{multirow}
\usepackage{hhline}

\usepackage[numbers]{natbib}

\title{Assessment of gait normality using a depth camera and mirrors}

\author{Trong-Nguyen Nguyen\\
DIRO, University of Montreal\\
Montreal, QC, Canada\\
\texttt{nguyetn@iro.umontreal.ca}\\
\And
Huu-Hung Huynh\\
University of Science and Technology\\
Danang, Vietnam\\
\texttt{hhhung@dut.udn.vn}\\
\And
Jean Meunier\\
DIRO, University of Montreal\\
Montreal, QC, Canada\\
\texttt{meunier@iro.umontreal.ca}\\
}

\iclrfinalcopy 
\begin{document}

\maketitle

\begin{abstract}
This paper presents an initial work on assessment of gait normality in which the human body motion is represented by a sequence of enhanced depth maps. The input data is provided by a system consisting of a Time-of-Flight (ToF) depth camera and two mirrors. This approach proposes two feature types to describe characteristics of localized points of interest and the level of posture symmetry. These two features are processed on a sequence of enhanced depth maps with the support of a sliding window to provide two corresponding scores. The gait assessment is finally performed based on a weighted combination of these two scores. The evaluation is performed by experimenting on 6 simulated abnormal gaits.
\end{abstract}

\section{Introduction}\label{sec:depthgaitintroduction}

This paper focuses on gait analysis using a system consisting of a Kinect 2, which employs the ToF depth estimation, and two mirrors to improve the body's depth map. When working with Kinect, the skeletal information is usually used for detecting abnormal gait (e.g.~\cite{Nguyen2016}). A limitation of this approach as well as any skeletal-based methods is that the skeleton could be deformed when there are self-occlusions of body parts in the depth image. Unfortunately, such deformation usually appears in pathological gaits. Therefore, an approach which can overcome this problem by employing other characteristics could be advantageous.

Inspired by the work~\cite{Yu2010} which showed that points of interest localized in a sequence of images could provide good results in classifying different human actions, we also try to determine key points using a simple detector. Instead of employing temporal information, our approach only uses a 2D detector on the depth image, then the gait assessment is performed with the support of a sliding window. Another difference is that the texton corresponding to each \textit{point of interest} (PoI) is not formed by a random forest. Instead, such texton is created by a processing flow of feature extraction, dimensionality reduction, and spatial discretization.

Besides, the \textit{level of posture symmetry} (LoPS) is also considered since pathological gaits usually occur with the appearance of asymmetric poses with a frontal view. Recent studies have also worked on such perspective to assess human gaits. In~\cite{Rougier2011}, the researchers measured gait symmetry using Depth Energy Image (DEI) which was computed as mean of depth silhouettes over gait cycles. Another approach with a similar goal was presented in~\cite{Moevus2014} employing the Multi-Dimensional Scaling (MDS) technique and shift invariant Euclidean distance. A common limitation of these works is that they depend on the result of gait cycle detection. In practical situations, it is difficult to determine cycles in pathological gaits since the movement of legs might be estimated with large deviation due to occlusions and/or strange postures (e.g.~\cite{Neurologic}). Therefore in our approach, the feature describing the level of symmetry is extracted based on the binary silhouette and a line separating each half-body.

A sequence of combinations of the two mentioned features is finally used to assess the gait normality. The remaining of this paper is organized as follow: the approach is presented in Section~\ref{sec:depthgaitapproach}, Section~\ref{sec:depthgaitexperiment} describes our experiments and evaluation, and the conclusion is shown in Section~\ref{sec:depthgaitconclusion}.

\section{Proposed approach}\label{sec:depthgaitapproach}

\begin{figure*}[!htb]
\centering
\scalebox{0.94}{
\begin{picture}(420,78)
	\put(0,0){\includegraphics[scale=0.714]{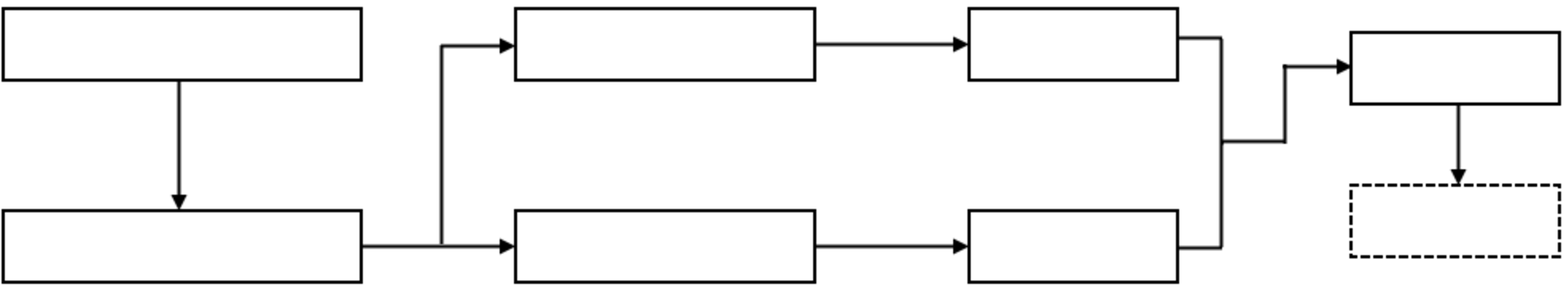}}
	\put(6,63){\small{Captured depth frames}}
	\put(6,8){\small{Enhanced depth frames}}
	\put(161,63){\small{PoI feature}}
	\put(157,8){\small{LoPS feature}}
	\put(274,63){\small{PoI-score}}
	\put(270,8){\small{LoPS-score}}
	\put(373,57){\small{Final score}}
	\put(378,16){\small{Decision}}
	\put(11,35){\small{unreliable point removal~\cite{NguyenKinect2}}}
	\put(228,64){\minibox{\small{model}\\\hspace{0.05cm}\small{based}}}
	\put(222,9){\minibox{\small{matching}\\\hspace{0.2cm}\small{based}}}
\end{picture}}
\caption{An overview of the proposed approach in this paper.}
\label{fig:depthgaitoverview}
\end{figure*}
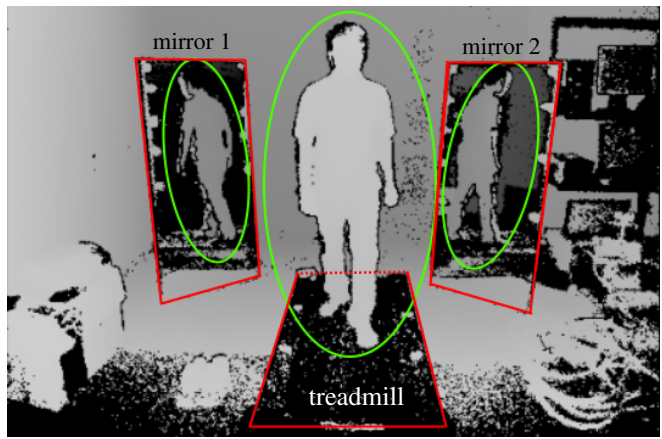
\begin{figure}[!htb]
	\centering
	\begin{picture}(250,163)
		\put(0,0){\includegraphics[width=0.62\textwidth]{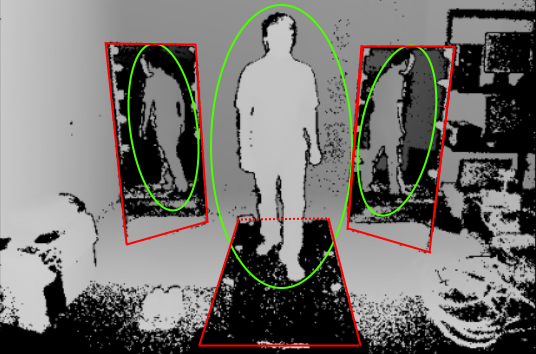}}
		\put(114,12){\textcolor[rgb]{1,1,1}{treadmill}}
		\put(55,147){\small mirror 1}
		\put(172,145){\small mirror 2}
	\end{picture}
	\caption{A raw depth map captured in our setup. This depth frame provides 3 collections of foreground pixels (highlighted by ellipses).}
	\label{fig:setup}
\end{figure}

An overview of the proposed approach is shown in Fig.~\ref{fig:depthgaitoverview}. Our setup consists of two mirrors, a treadmill and a ToF depth camera to capture a frontal view of the subject. A typical raw depth map is displayed in Fig.~\ref{fig:setup}. The input of the processing flow is a sequence of enhanced (frontal view) depth maps of the subject walking on the treadmill. These enhanced depth maps are formed by a combination of 3 parts of the raw depth map (see Fig.~\ref{fig:setup}) corresponding to the real body on the treadmill and two additional views provided by the two mirrors. These parts are highlighted by 3 ellipses in Fig.~\ref{fig:setup}. A point appearing in a mirror is simply reflected through it to obtain its true 3D position~\cite{NguyenKinect2}. This combination of depth information improves the quality of the depth map of the subject. In addition, some pixels inside a mirror are not considered due to depth measurement ambiguities which occur from unwanted multiple reflections~\cite{Freedman2014, NguyenKinect2}. The final result is an enhanced frontal view depth map of the subject walking on a treadmill.

\subsection{PoI-based feature}\label{sec:depthgaitPoIfeature}

\begin{figure}[!htb]
\centering
\begin{picture}(200,160)
	\put(60,33){\includegraphics[scale=0.75]{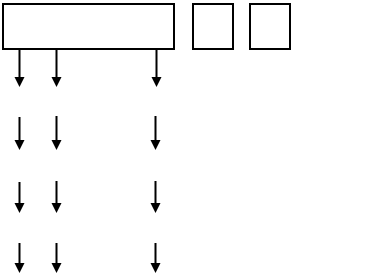}}
	\put(37,145){\LARGE{.....}}\put(195,145){\LARGE{.....}}
	\put(-20,142){Depth frame} \put(84,142){$frame_t$}
	\put(-20,108){Key point} \put(64,108){$P_1$} \put(81,108){$P_2$} \put(98,110){.........} \put(125,108){$P_n$}
	\put(-20,79){Raw feature} \put(64,79){$f_1$} \put(81,79){$f_2$} \put(98,81){.........} \put(126,79){$f_n$}
	\put(-20,52){3--d vector} \put(64,52){$v_1$} \put(81,52){$v_2$} \put(98,54){.........} \put(126,52){$v_n$}
	\put(-20,25){1--d index} \put(64,25){$i_1$} \put(81,25){$i_2$} \put(98,27){.........} \put(127,25){$i_n$}
	\put(62,14){\includegraphics[height=0.3cm,width=2.6cm]{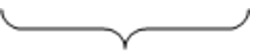}}
	\put(142,14){\includegraphics[height=0.3cm,width=0.8cm]{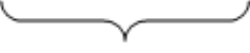}}
	\put(172,14){\includegraphics[height=0.3cm,width=0.8cm]{depthgait/bracket}}
	\put(-20,02){Histogram} \put(95,2){$h_t$} \put(145,2){$h_{t+1}$} \put(174,2){$h_{t+2}$}
	\put(-23,-2){\framebox(250,162)}
\end{picture}
\caption{Steps of extracting a histogram describing an enhanced depth map.}
\label{fig:depthgaitPoI}
\end{figure}

One of the contributions in our approach is the process of describing each enhanced depth map by a histogram formed by a collection of PoI. An overview of this stage is shown in Fig.~\ref{fig:depthgaitPoI}. Concretely, key points are localized based on each depth frame given a 2D detector together with an additional criterion. A raw feature is then calculated for each key point before reducing the number of dimensions to 3. In order to form a histogram representing the considered frame, the 3-dimension vector of each PoI is quantized to give an index. 


\subsubsection{Key point localization}
\begin{figure}[!htb]
	\centering
	\begin{picture}(190,100)
		\put(0,0){\includegraphics[scale=0.75]{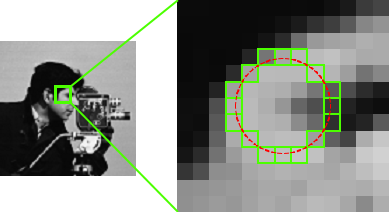}}
		\put(126,46){\textcolor[rgb]{1,1,1}{$p$}}
	\end{picture}
	\caption{An example of a pixel $p$ and 16 contiguous ones in the Bresenham circle of radius 3.}
	\label{fig:depthgaitFASTpoint}
\end{figure}

The FAST (Features from Accelerated Segment Test) detector~\cite{Rosten2006} is employed in our approach because it is not time-consuming. In detail, a pixel $p$ is assigned to be a PoI if the depth values of the 16 contiguous pixels in the Bresenham circle of radius 3 centered at $p$ are all greater or less than $p$'s depth with a predefined threshold (see Fig.~\ref{fig:depthgaitFASTpoint}). A pixel $p$ is discarded by the detector if itself or any of the 16 contiguous pixels is in the background.

\subsubsection{Raw feature extraction}

Assume that there are $n$ key points localized in a depth frame $frame_t$. Each point of interest $P_k$ $(1 \le k \le n)$ is represented by its coordinates $P^{3d}_k$ together with 16 related ones $P^{3d}_{k,j}$ $(1 \le j \le 16)$ in the 3D space. A simple feature describing such point is formed by combining 16 vectors pointing from the point $P_k$
\begin{equation}
	f_k = \Psi_{j=1}^{16}(P^{3d}_{k,j}-P^{3d}_k)
	\label{eq:depthgaitrawfeature}
\end{equation}
where $\Psi(\cdot)$ denotes a function of concatenation. The result of the $\Psi$ function indicates the relation between a PoI and the 16 points around. However, the position of the PoI related to the whole body should also be considered since similar features $f_k$ may occur at different body parts. Therefore the ratio of the distance between the PoI and the ground to the body height is added to the result of (\ref{eq:depthgaitrawfeature}). The obtained feature thus has 49 dimensions.

\subsubsection{Dimensionality reduction}

In order to reduce redundancies in the calculated feature, Principal Component Analysis (PCA) is applied to provide a new representation $v_k$ of $f_k$ (see Fig.~\ref{fig:depthgaitPoI}) with a significantly lower length. The advantages of PCA are its simplicity, good efficiency in many applications, and the projection of new samples based on a matrix estimated during the training stage. In our approach, the target dimension is 3.

\subsubsection{Histogram formation}\label{sec:depthgaithistogram}

As mentioned at the beginning of Section~\ref{sec:depthgaitapproach}, each depth frame needs to be represented by a histogram of detected key points. Inspired by the work~\cite{Mignotte2008}, a spatial quantization is performed on the dimensional-reduced space (3D in our work). Its range is determined by the minimum and maximum in each dimension. The index of a 3-dimension vector $v_k$ in Fig.~\ref{fig:depthgaitPoI} and the value of the corresponding bin in the histogram are respectively estimated and updated as
\begin{equation}
	i_k= \sum_{j=1}^3 q^{j-1} \lfloor q(v_{k,j}-min_j)(max_j-min_j+ \epsilon )^{-1} \rfloor
	\label{eq:depthgaitindex}
\end{equation}
\begin{equation}
	h_t[i_k] \leftarrow h_t[i_k] + 1
	\label{eq:depthgaithistogram}
\end{equation}
where $q$ is the number of bins in each dimension, $\epsilon$ is a small value which guarantees that $i_k$ falls in the range $[0, q^3-1]$, $min_j$ and $max_j$ are respectively minimum and maximum values in the $j^{th}$ dimension. The calculated histogram thus has $q^3$ bins, $q$ was set to 5 in our experiment. Notice that the histogram $h_t$ is not normalized because the number of detected key points can be a characteristic supporting gait assessment.
The collection of PoI is thus represented by a histogram computed for each depth frame.

\subsection{LoPS features}
\begin{figure*}[htb]
\centering
\scalebox{0.95}{
\begin{picture}(404,180)
	\put(10,20){\includegraphics[scale=0.9]{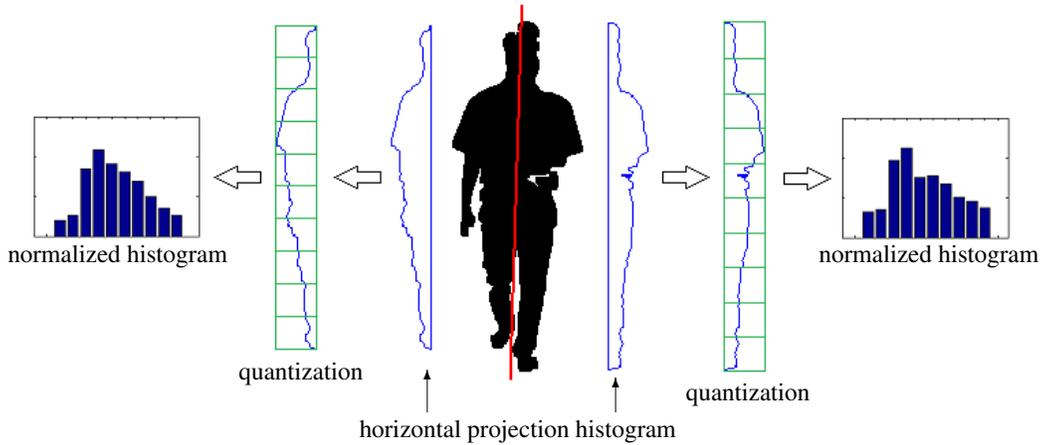}}
	\put(140,0){horizontal projection histogram} \put(167,10){\vector(0,1){17}}\put(242,10){\vector(0,1){14}}
	\put(92,23){quantization} \put(270,15){quantization}
	\put(0,70){normalized histogram} \put(323,70){normalized histogram}
\end{picture}}
\caption{An example of a body silhouette with the corresponding separation line, two histograms of horizontal projection, and results after quantization of 10 bins. The half-body ratios are respectively 0.48 and 0.52 from left to right.}
\label{fig:depthgaithalfprojection}
\end{figure*}

As mentioned in Section~\ref{sec:depthgaitintroduction}, gait anomalies often appear via gait asymmetry in frontal view (e.g.~\cite{Bauckhage2009}). In this section, a statistical feature describing level of posture symmetry is presented, in which the considered data is the sequence of binary silhouettes corresponding to the body in frontal view. The idea of this process is that the body silhouette is decomposed into left and right half-body parts using a separation line, and each one is represented by an ensemble of characteristics. In each frame, this line is formed by the pixel indicating the head and its projection onto the ground, in which the ground parameters are determined by the calibration, and the head location is provided by a built-in function of the used device (Microsoft Kinect in our work). The LoPS features are:

\subsubsection{Half-body ratio}\label{sec:depthgaitratio}
This ratio is simply given by
\begin{equation}
	\begin{cases}ratio_{left} = \#(pixel_{left})/\#(pixel_{whole})\\ratio_{right} = 1 - ratio_{left}\end{cases} 
	\label{eq:depthgaitratio}
\end{equation}
where $\#(pixel_{left})$ and $\#(pixel_{whole})$ indicate the number of pixels in left half-body and whole body silhouette, respectively. Although this characteristic seems quite simple, it gives a good description since in the symmetric gait, i.e. gait with normal postures, the difference between these two ratios is small compared with asymmetric ones.

\subsubsection{Histogram of horizontal projection}\label{sec:depthgaithistprojection}
This feature describes the vertical distribution of half-body pixels. As illustrated in Fig.~\ref{fig:depthgaithalfprojection}, each half-body is processed to extract the number of pixels lying in each row. The result when considering each frame is thus a pair of histograms with different (sometimes equivalent) number of bins. In order to perform a comparison, these histograms are quantized into a fixed number of bins. The employed quantization technique is inspired from~\cite{Bauckhage2009}. The results are finally normalized, i.e. these histograms are represented by unit vectors.

\subsection{Gait assessment}

As described above, the stage of feature extraction provides a PoI histogram (Section~\ref{sec:depthgaitPoIfeature}), a pair of half-body ratios (Section~\ref{sec:depthgaitratio}), and two normalized histograms (Section~\ref{sec:depthgaithistprojection}) for each depth frame. The gait assessment is then performed based on a sequence of consecutive depth frames. Given a sequence of the mentioned features, two local scores are individually estimated for PoI and LoPS.

\subsubsection{PoI-score}\label{sec:depthgaitPoIscore}

Inspired by the work~\cite{Nguyen2016}, this process tries to build a model of normal gait describing the transition of posture characteristics. Then the log-likelihood value provided by the trained model is used as the PoI-score. Because extracting gait cycles as in~\cite{Nguyen2016} is complicated since the skeletal information does not exist in our approach, we model sequences of depth maps based on the support of a sliding window with a predefined width. The hidden Markov model (HMM) technique is employed because of its strength in human gait related problems (e.g.~\cite{Nguyen2016,Tao2016}). A HMM with multi-dimensional observations has a high computational cost, thus a modification is applied on the input, i.e. a sequence of histograms, to give scalar observations. In detail, this stage tries to model the change of distance between two consecutive histograms. It means that the histogram at time $t$, $h_t$, is converted into a scalar value $\Delta h_t$ as
\begin{equation}
	\Delta h_t = H(h_t,h_{t-1})
\end{equation}
where $H(\cdot,\cdot)$ denotes a function estimating the Hamming distance~\cite{Norouzi2012} between two arguments. Instead of employing a HMM with specific transitions as in~\cite{Nguyen2016}, a HMM with full connection is used since it models the change of frames in a sliding window rather than a gait cycle. Besides, the observations are modeled by a Gaussian Mixture Model (GMM) since they are represented by a continuous variable.

\subsubsection{LoPS-score}\label{sec:depthgaitLoPSscore}

The goal of this score is to give a measurement of similarity between left and right parts of a body silhouette which are separated by a line connecting the head and its projection onto the ground. Since this score is measured on a sliding window, each of the two body parts contains a sequence of half-body ratios and a sequence of normalized histograms, that could be considered as a temporal pattern of wave form. In an ideal case, i.e. the gait is absolutely symmetric, the only difference between two body parts is the phase, thus a good matching would be obtained after performing an appropriate shifting. Therefore we estimate a matching-based measurement between the two sequences of ratios as well as the two sequences of histograms to determine the level of symmetry.
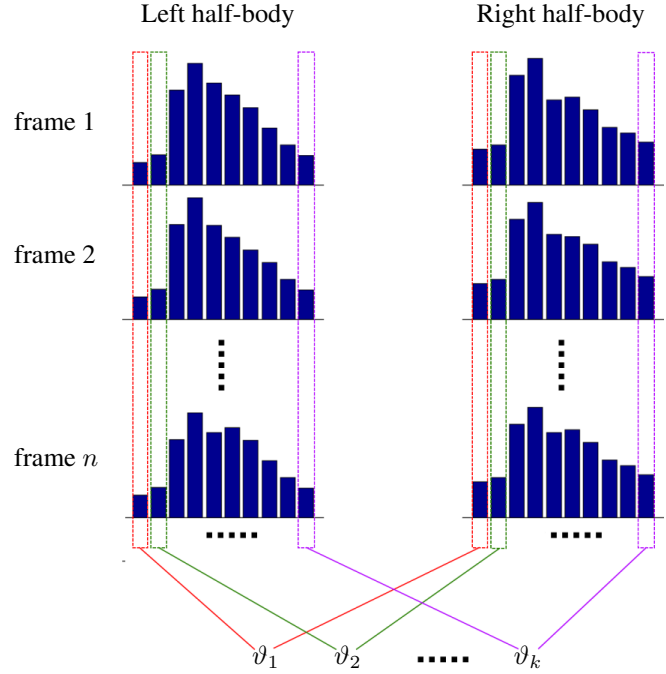
\begin{figure}[!htb]
\centering
\begin{picture}(260,255)
	\put(40,0){\includegraphics[scale=0.47]{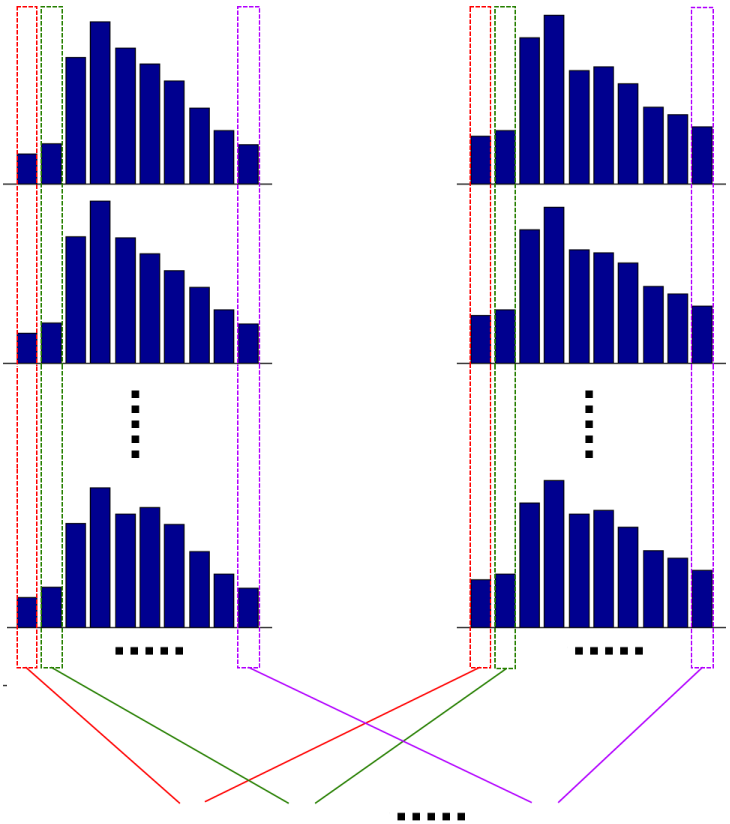}}
	\put(48,243){Left half-body}
	\put(175,243){Right half-body}
	\put(0,202){frame 1}
	\put(0,152){frame 2}
	\put(0,75){frame $n$}
	\put(90,0){$\vartheta_1$}
	\put(121,0){$\vartheta_2$}
	\put(189,0){$\vartheta_k$}
\end{picture}
\caption{Similarity measurement between two sequences of $n$ consecutive histograms corresponding to two parts of a body silhouette. The returned value is computed as mean of $k$ measured cross-correlations $\vartheta$.}
\label{fig:depthgaithistmatching}
\end{figure}
Let us consider two sequences of half-body ratios. By applying the cross-correlation technique~\cite{Stoica2005}, which measures the similarity between a sequence and shifted copies of the other, the returned value is assigned as the maximum of calculated matching similarities. This process is more complicated when working on sequences of histograms, the cross-correlation is thus applied on $k$ pairs of sequences corresponding to the $k$ bins of projection histograms. A visualization of this calculation is shown in Fig.~\ref{fig:depthgaithistmatching}. In summary, the LoPS-score is estimated as
\begin{equation}
	S_{LoPS} = log\big[\frac{1}{2} \big(\zeta_r + \frac{1}{k} \sum_{i=1}^k \vartheta_i \big)\big]
\label{eq:depthgaitLoPSscore}
\end{equation}
where $\zeta _r$ is the score of similarity between two sequences of half-body ratios, $k$ is the number of bins of each quantized horizontal projection histogram, and $\vartheta_i$ is the measure of similarity between two sequences of values corresponding to the $i^{th}$ bin (see Fig.~\ref{fig:depthgaithistmatching}).

\subsubsection{Score combination}
In our approach, the human gait in a sequence of depth frames is assessed by a score which is calculated based on the PoI-score and LoPS-score. A weighted combination is employed where the weights are estimated in the training process. Given a predefined width $w$ of the sliding window, a sequence of $n$ depth frames could provide $l = (n-w+1)$ overlapped sequences of $w$ consecutive frames corresponding to $l$ inputs for the computation of PoI- and LoPS-scores. By performing the computations described in Section~\ref{sec:depthgaitPoIscore} and~\ref{sec:depthgaitLoPSscore}, we obtain $l$ pairs of scores. Let us notice that the likelihood provided by a HMM is in the range [0, 1], $\zeta _r$ and $\vartheta_i$ in (\ref{eq:depthgaitLoPSscore}) also fall in this range because of the cross correlation, the two scores are thus negative since a logarithm function was applied. Because the training dataset consists of normal gait patterns, these scores are expected to be near the value of 0. Therefore, the weight of the score with lower absolute value should be stronger. In summary, the two weights are calculated based on $l$ pairs of scores as follows
\begin{equation}
	 \begin{cases} w_{LoPS} = \big(\sum_{i=1}^{l} S_{PoI}^{(i)}\big)/ \big(\sum_{i=1}^{l} S_{PoI}^{(i)} + \sum_{i=1}^{l} S_{LoPS}^{(i)}\big) \\w_{PoI} = 1-w_{LoPS}\end{cases} 
\label{eq:depthgaitcombination}
\end{equation}
where $\big(S_{PoI}^{(i)}, S_{LoPS}^{(i)}\big)$ is the pair of scores computed from the $i^{th}$ training sequence of depth frames. Finally, the gait in a new sequence of depth frames is assessed based on a comparison between a predefined threshold and the final score which is estimated by a weighted sum as
\begin{equation}
	S = w_{PoI}.S_{PoI} + w_{LoPS}.S_{LoPS}
\label{eq:depthgaitfinalscore}
\end{equation}

\section{Experiment}\label{sec:depthgaitexperiment}
\begin{table*}[tp]
\centering
\caption{Classification errors corresponding to our system and a modification of method~\cite{Bauckhage2009}.}
\label{table:depthgaitresult}
\vspace{5pt}
\begin{tabular}{|l||c|c|c|c|c|}
\hline
Evaluation   & PoI  & LoPS & PoI + LoPS & One-class SVM~\cite{Bauckhage2009} & Binary SVM~\cite{Bauckhage2009} \\ \hline \hline
per-frame    & 0.47 & 0.32 & 0.31        & 0.14                   & 0.11                \\ \hline
per-sequence & 0.56 & 0.06 & \textbf{0.06}        & 0.19                   & 0.14                \\ \hline
\end{tabular}
\end{table*}

Our experiments were performed with normal gaits and 6 simulated abnormal ones which were created by padding a sole of different thicknesses (10, 15 and 20 cm) under the left or right foot. The data of each walking gait were acquired as 1200 depth frames with a frame rate of 13 fps.

The dataset was collected from 9 volunteers. We employed the gaits of 6 subjects as the training data, and the gaits of the 3 remaining subjects were the test set. In our experiments, the threshold of FAST detector was 30 millimeters, the value of $q$ in (\ref{eq:depthgaitindex}) was 5, the number of HMM hidden states was 8 and the width of the sliding window was 10 corresponding to 0.77 seconds. Histograms of horizontal projection were split into 10 bins. These values were selected according to our experience on previous gait-related works. 
In order to provide a comparison, the method~\cite{Bauckhage2009} was considered since it also extracts features from frontal view subject's silhouettes and performs the classification on a sequence of frames. Beside the original binary SVM proposed in~\cite{Bauckhage2009}, we slightly modified it to create a one-class SVM because we should consider the case that the available data contain only normal gaits. In other words, this modification is reasonable since there are numerous abnormal gait types in practical situations, collecting such gait samples with a high generalization would be difficult.

In the training stage, our model and the one-class SVM were trained using only normal gaits of 6 subjects in the training set, while the binary SVM employed both normal and abnormal ones. The testing stage was performed on the test set of the 3 remaining subjects. Similarly to related works on binary decision (e.g.~\cite{Nguyen2016,Tao2016}), we used a measure calculated from the Receiver Operating Characteristic (ROC) curve to evaluate our system. The Equal Error Rate (EER) is employed since it represents a trade-off between False Positive Rate (FPR) and False Negative Rate (FNR) and is comparable with typical classification errors.

Table~\ref{table:depthgaitresult} shows the errors resulting from our system and the two models of~\cite{Bauckhage2009}. The term \textit{per-frame} indicates frame-based assessment, in which the gait normality of each frame is estimated based on 10 (i.e. width of sliding window) and 21 recent frames for our method and the two models of~\cite{Bauckhage2009}, respectively. The per-sequence assessment was performed on each walking gait (1200 frames) using average score in our work and a trigger of 30 consecutive frames in~\cite{Bauckhage2009}. The classification errors using only the PoI were significantly larger compared with the LoPS. This explains to some extent why the trained weights in (\ref{eq:depthgaitfinalscore}) were 0.0014 and 0.9986 for the PoI and LoPS, respectively. However, the combination of these two features provided better results. It other words, the confidence of a score output from our system increases when employing the two proposed features instead of only one. We can say that the LoPS feature plays the main role in our system, and the PoI performs a tuning on the output. Table~\ref{table:depthgaitresult} also shows that our approach outperformed the work~\cite{Bauckhage2009} when performing the assessment on full sequences of walking gaits. The ability of assessing gait balance in our system is thus better than~\cite{Bauckhage2009} when combining posture assessments over a sequence. In other words, the sequence assessment in~\cite{Bauckhage2009} is significantly easier to be affected by noise compared with our method. Although both works focused on the gait balance embedded in the subject's silhouette, the LoPS aims to measure the gait symmetry while the lattice-based feature in~\cite{Bauckhage2009} attempts to distinguish between normal gait and high-imbalance ones such as wavering, faltering, and falling.

\section{Conclusion}\label{sec:depthgaitconclusion}
This paper presents an original work that employs enhanced depth maps provided by a depth camera with the help of mirrors to assess human gait normality. The assessment is performed using a combination of a PoI-score, which is computed based on key points in a sequence of depth frames, and a LoPS-score describing a measurement of body balance. In this system, the LoPS feature is the main factor to estimate the score measuring the gait normality, and the PoI performs a tuning on the resulted score. The ability of our system has been demonstrated via the experiments on 6 simulated abnormal gaits. In order to improve this approach, further works will be carried on including (1) proposing other PoI-based features, (2) trying other values of parameters (e.g. number of bins of quantized horizontal projection histogram), (3) investigating full-body 3D key point detectors, and (4) collecting a larger dataset.

\subsubsection*{Acknowledgement}
The authors would like to thank the NSERC (Natural Sciences and Engineering Research Council of Canada) for supporting this work (Discovery Grant RGPIN-2015-05671). This work was also supported by The Ministry of Education and Training, Vietnam, Grant KYTH-59.

\bibliography{references}
\bibliographystyle{iclr2019_conference}

\end{document}